\documentclass{article}

     \PassOptionsToPackage{numbers, compress}{natbib}

     \usepackage[preprint]{neurips_2021}

\usepackage[utf8]{inputenc} 
\usepackage[T1]{fontenc}    
\usepackage{hyperref}       
\usepackage{url}            
\usepackage{booktabs}       
\usepackage{amsfonts,amsthm,amsmath,amssymb}       
\usepackage{nicefrac}       
\usepackage{microtype}      
\usepackage{xcolor}         
\usepackage{float}
\usepackage{graphicx}

\theoremstyle{plain}
\newtheorem{theorem}{Theorem}[section]

\newtheorem{algorithm}[theorem]{Algorithm}
\theoremstyle{definition}

\title{Design equivariant neural networks for 3D point cloud}

\author{%
  Thuan N.A. Trang\\
  FPT Software Ho Chi Minh\\
  Ho Chi Minh City, Vietnam\\
  \texttt{thuantna@fsoft.com.vn} \\
  \And
  Thieu N. Vo\\
  Ton Duc Thang University and FPT Software\\
  Ho Chi Minh City, Vietnam\\
  \texttt{vongocthieu@tdtu.edu.vn} \\  
  \And
  Khuong D. Nguyen\\
  FPT Software Ho Chi Minh\\
  Ho Chi Minh City, Vietnam\\
  \texttt{khuongnd6@fsoft.com.vn}
}

\begin{document}

\maketitle

\begin{abstract}
This work seeks to improve the generalization and robustness of existing neural networks for 3D point clouds by inducing group equivariance under general group transformations.
The main challenge when designing equivariant models for point clouds is how to trade-off the performance of the model and the complexity.
Existing equivariant models are either too complicate to implement or very high complexity.
The main aim of this study is to build a general procedure to introduce group equivariant property to SOTA models for 3D point clouds.
The group equivariant models built form our procedure are simple to implement, less complexity in comparison with the existing ones, and they preserve the strengths of the original SOTA backbone.
From the results of the experiments on object classification, it is shown that our methods are superior to other group equivariant models in performance and complexity. 
Moreover, our method also helps to improve the mIoU of semantic segmentation models. 
Overall, by using a combination of only-finite-rotation equivariance and augmentation, our models can outperform existing full $SO(3)$-equivariance models with much cheaper complexity and GPU memory.
The proposed procedure is general and forms a fundamental approach to group equivariant neural networks.
We believe that it can be easily adapted to other SOTA models in the future.
\end{abstract}

\section{Introduction}
The unprecedented development in 3D acquisition technologies has led to the widely affordable of 3D sensors such as LiDARs, and RGB-D cameras, which directly
output data termed Point Cloud. Theoretically, this data can provide rich information
in terms of geometric, shape, and scale. However, in practice, capturing the same object from different locations or angles produces different data. Consequently, inducing equivariance under transformations such as permutation, translation and rotation in deep neural network architectures is crucial to improving generalization when dealing with 3D point clouds.

Incorporating equivariant properties in deep neural network architectures has long been a powerful idea. For example, Convolutional Neural Networks (CNNs) can effectively extracting local information in 2D image by having translation equivariance. Therefore, general theory for group equivariant neural networks and their universality have been recently developed. Some research effort was put to study 2D and 3D rotation equivariance for grid or voxel images with positive results. However, few attention has been paid to the study of group equivariant neural networks for 3D point clouds. 

This work seeks to improve generalization and data efficiency of neural networks for 3D point clouds through group equivariance since processing raw point cloud directly can eliminate redundant computation and memory required. In particular, we propose a general procedure, called $G$-PointX, to introduce group equivariant to an existing SOTA backbone which is not equivariant yet.
The main idea of the procedure is inherited by the symmetricization technique, called Reynolds operator (see \cite{sturmfels2008algorithms}).
For practical applications, we apply this procedure to two typical backbones which are PointNet++ and PointConv to obtain efficient group equivariant models for 3D point clouds that we term $G$-PointNet and $G$-PointConv, respectively.

We demonstrate the effectiveness of $G$-PointX through extensive empirical experiments. For the classification benchmark, $G$-PointX outperforms state-of-the-art deep neural networks for 3D point clouds when dealing with rotated data. 
For semantic segmentation benchmark, $G$-PointX outperforms original models by a significant margin. 
In short, the main contribution of this paper are:
\begin{itemize}
	\item A detailed formulation of a novel group equivariant CNN and MLP termed $G$-PointConv and $G$-PointNet++
	\item Comprehensive and extensive experiments demonstrating the effectiveness of our proposed method.
\end{itemize}

\section{Related works}
\textbf{Deep learning with raw 3D point clouds:} Recently, several approaches extracting features from 3D point clouds have been presented in the literature. As a pioneer in raw point cloud processing, PointNet \cite{qi2017pointnet} extracted features by simply employing a combination of MLPs and max-pooling. PointNet++ \cite{qi2017pointnet++}, afterwards, introduces a new architecture called hierarchical structure which efficiently aggregated information from local area. Notably, PointNet++ used PointNet operation to extract feature, which could be considered as MLP operation for point cloud. Later papers inspired by PointNet++ often keep the hierarchical structure as the same and only modify the operation. Typically, PointConv\cite{wu2019pointconv} replaced MLPs with Convolution layers and PointTransformer\cite{zhao2021point} used self-attention operation to collect information in local region. As only relied on relative position and color, PointNet++, PointConv\cite{wu2019pointconv} and PointTransformer\cite{zhao2021point} are equivariant with translation and permutation.  
	
\noindent
\textbf{Group equivariant neural networks:} There are several methods tried to embed group equivariant properties into the architecture. For example, the patch-wise $360^o$-rotation equivariance in Harmonic Networks \cite{worrall2017harmonic} was obtained by using circular harmonics. Moreover, \cite{cohen2016steerable} and \cite{jacobsen2017dynamic} proposed blocks named Steerable CNNs, which can be equivariant to group $SO(3)$ of 3D rotations. This type of block, then, applied on capsule networks \cite{sabour2017dynamic} and with N-body networks \cite{kondor2018n}. The notions of equivariance and convolution were also generalized in neural networks to the action of compact groups in \cite{kondor2018generalization}. Apart from that, some further abstract setting and universality results on group equivariance are described in \cite{kumagai2020universal, maron2019universality, petersen2020equivalence, ravanbakhsh2020universal, yarotsky2018universal}.

\noindent
\textbf{Group equivariant neural networks for 3D point clouds:} Despite the rapid development in group equivariant neural networks, the number of papers designing group equivariant neural networks for point cloud is modest. However, rotation equivariance seems indispensable for architecture in this field since the point clouds are heavily depend on the location and the orientations of the Lidar sensors. In previous works, that property was achieved by several approaches. For instance, while \cite{thomas2018tensor} used filters built from spherical harmonics, \cite{chen2019clusternet, li2020rotation, zhang2019rotation} employed the rigorous rotation invariance representation of point clouds in terms of angles and distances. Furthermore, we can also apply a multi-level abstraction involving graph convolutional neural networks \cite{Kim2020advances} or quaternion-based neural networks \cite{shen20193d, zhang2020quaternion, zhao2020quaternion} to acquire rotation equivariance.
		
\noindent
\textbf{Our work stands out from other approaches:} 
The method we use in $G$-PointX is different from the previous group equivariant models. Rather than design new representation or architecture, we provide a plug-in to existing models, which allows us to inherit the advantages of the original model. In our method, we achieve group equivariance by applying the standard MLPs and CNNs, and re-arranging different group conjugations of the operations in a suitable way. 

\section{Groups and group actions}


Let $G$ be a group with identity $1$ and $V$ a nonempty set. 
An action of $G$ to $V$ is a map $G \times V \to V$ defined by a certain operation $(g,v) \mapsto gv$ satisfying the properties $1v=v$ and
$$(g_1g_2)v = g_1(g_2v), \quad \forall g_1,g_2 \in G, \, \forall v \in V.$$ 
If $G$ acts on $V$, then $G$ acts on every objects built on $V$.
For example, if $G$ acts on $\mathbb{R}^3$, 
then $G$ acts naturally on functions defined on $\mathbb{R}^3$, the space $\mathcal{X}$ of point clouds on $\mathbb{R}^3$, and the functions defined on $\mathcal{X}$.
The following two group actions are important in our consideration:
\begin{itemize}
	\item $G = S_N$ is the permutation group of $\{1,2,\ldots,N\}$, and $G$ acts on a point cloud containing $N$ points in $\mathbb{R}^3$ by permuting the arrangement of points.
	
	\item $G$ is a subgroup of the group $SO(3)$ of 3D rotations, and $G$ acts on $\mathbb{R}^3$ by matrix-vector multiplication.
	
	\item The group $\mathbb{R}^3 \rtimes G$ which is the semiproduct of the transition group $\mathbb{R}^3$ and a subgroup $G$ of $SO(3)$, and this group acts on $\mathbb{R}^3$ by $(q,h)x = q+hx$ for $(q,h) \in \mathbb{R}^3 \rtimes G$ and $x \in \mathbb{R}^3$.
\end{itemize}

It is noted that the group multiplication in $\mathbb{R}^3 \rtimes G$ is determined as $(q,h) \cdot (p,g) = (q+hp,hg)$ for some $(p,g),\,(q,h) \in \mathbb{R}^3 \rtimes G$. Thus
$(q,h)^{-1} = (-h^{-1}q,h^{-1})$
and 
$(q,h)^{-1} \cdot (p,g) = (h^{-1}(p-q),h^{-1}g)$.


Now we consider a 3D point cloud as a function from a finite set $\mathcal{P} \subset \mathbb{R}^3$ to $\mathbb{R}^d$, where $d$ is the dimension of the feature vectors.
Sometimes, we extend the domain of this function from $\mathcal{P}$ to the whole $\mathbb{R}^3$ and view each point cloud as a continuous function (with compact support) from $\mathbb{R}^3$ to $\mathbb{R}^d$ in order to involve existing strong techniques in mathematics in developing suitable FFNNs for point clouds. 
A point cloud will be then identified with a function on $\mathcal{C}(\mathcal{P},\mathbb{R}^d)$ or $\mathcal{C}(\mathbb{R}^3,\mathbb{R}^d)$. 

An FFNN is defined to be a sequence
$$\Phi_1 \mapsto \sigma \mapsto \Phi_2 \mapsto \sigma \mapsto \ldots \mapsto \sigma \mapsto \Phi_L$$
of transformations, where each $\Phi_i$ is a linear or nonlinear transformation 
between the $i$-th layer $\mathcal{C}(\mathbb{\mathcal{P}}^{(i)},\mathbb{R}^{r_i})$ to the $(i+1)$-th layer $\mathcal{C}(\mathbb{\mathcal{P}}^{(i+1)},\mathbb{R}^{r_{i+1}})$ 
followed by the nonlinear point-wise activation function $\sigma$.
MLPs and CNNs are two types of operators that considered in this work, which are the most used FFNNs in practice.


Assume that $G$ acts on $\mathbb{R}^3$.
Then $G$ also acts on functions on point clouds $L=\mathcal{C}(\mathbb{R}^3,\mathbb{R}^d)$ as follows:
for each $g \in G$ and $f \in L$, the action $gf \in L$ is defined by
$$[gf](x) = f(g^{-1} x), \quad x \in \mathbb{R}^3.$$ 
We can describe the action of $G$ on $L$ intuitively as follows: when we rotate the image $f$ due to the orientation $g$, then the feature vector of the new image at the coordinate $x$ is exactly the feature vector of the old image at the coordinate $g^{-1} x$.

Let $\Phi:\mathcal{C}(\mathbb{R}^{3},\mathbb{R}^{d_1}) \to \mathcal{C}(\mathbb{R}^{3},\mathbb{R}^{d_2})$ be a linear (or nonlinear) transformation. 
We say that $\Phi$ is $G$-equivariant if and only if  $\Phi(gf)=[g\Phi](f)$ for every $g \in G$ and $f \in \mathcal{C}(\mathbb{R}^{3},\mathbb{R}^{d_1})$.
An FFNN is called $G$-equivariant if all of its transformations are $G$-equivariant.

Thanks to the point-wise structure, every nonlinear activation function is $G$-equivariant with respect to any group $G$ (see \cite{cohen2016group, kondor2018generalization}). 
However, not every linear transformation is $G$-equivariant.
Therefore, in the next section we give a natural approach to refine a given FFNN and produce an equivariant one.
The notable  PointNet++\cite{qi2017pointnet++} and PointConv\cite{wu2019pointconv} layers with rotation groups $G$ will be considered to illustrate the theory and to test the effectiveness of the proposed approach.



\section{A general scheme for constructing $G$-equivariant NNs for 3D point clouds}

\subsection{Local feature extractions}

Given a point cloud $\mathcal{X} = \left\{ (p,f_p) \right\}_{p \in \mathcal{P}}$ on $\mathbb{R}^3$, 
most of the state-of-the-art models for machine learning tasks on 3D point clouds are built based on a local feature extraction.
In general, a local feature extraction extracts important information from a group of points around a given point and it is formulated by 
\begin{equation}\label{eq:local_feature_extraction}
f'_q = F(\{p-q, f_{p},f_{q}\}_{p \in \mathcal{N}(q)}),
\end{equation}
where 
$q$ is the centroid, 
$\mathcal{N}(q)$ is the local neighbor of $q$
and $F$ is a local aggregration.

In this paper, we consider the aggregation methods in PointConv and PointNet++, whose formula can be written as:
\begin{itemize}
    \item PointNet++
    $$f'_q = \text{MLP}_1 \left(
    \mathcal{R}
    \left\{ 
    \text{MLP}_2 \left( (p_i-q) \oplus f_{p_i} \right),\,{p_i \in \mathcal{N}(q)}
    \right\}
    \right),$$
    
    \item PointConv
    $$
    f'_q = \sum\limits_{p_i \in \mathcal{N}(q)}
    \text{MLP}_1(||p_i-q||) \cdot \text{MLP}_2(p_i-q) \cdot f_{p_i}.
    $$
\end{itemize}

\subsection{A general framework for introducing equivariance}


For the sake of simplicity, we choose $G$ as a finite group. In case the transformation group is infinite, a finite subgroup is chosen. We will construct a neural networks of the form of a sequence of transformations
\begin{align*}
\mathcal{C}(\mathcal{P}^{(0)} \times G,\mathbb{R}^{d_{0}})
\xrightarrow{\Phi^{(1)}}
\mathcal{C}(\mathcal{P}^{(1)} \times G,\mathbb{R}^{d_{1}})
\xrightarrow{\Phi^{(2)}}
\ldots
\xrightarrow{\Phi^{(L)}}
\mathcal{C}(\mathcal{P}^{(L)} \times G,\mathbb{R}^{d_{L}}).
\end{align*}
Here, 
\begin{itemize}
	\item $\mathcal{P}^{(0)} = \mathcal{P}$ is the initial point cloud.
	\item $\mathcal{P}^{(l)}$ is a subset of $\mathcal{P}^{(l-1)}$ and it is chosen by using the furthest point sampling algorithm.
	\item Each grouping layer $\Phi^{(l)}$ is a $G$-equivariant transformation.
\end{itemize}

For each $l=1,\ldots,L$, the grouping layer $\Phi^{(l)}$ maps a point clouds $\mathcal{X} = \{(p,g),f^{(l-1)}_{(p,g)}\}_{(p,g) \in \mathcal{P}^{(l-1)} \times G}$ to a point cloud $\mathcal{Y} = \{(q,h),f^{(l)}_{(q,h)}\}_{(q,h) \in \mathcal{P}^{(l)} \times G}$.
Here, $\mathcal{P}^{(l)}$ is a subset of $\mathcal{P}^{(l-1)}$ and is chosen by using the farthest point sampling algorithm.
For each point $(q,h)$ in $\mathcal{P}^{(l)} \times G$, the feature vector $f^{(l)}_{(q,h)}$ is determined as
\begin{align}\label{eq:group_layer_V3_h}
f^{(l)}_{(q,h)} = \tilde{F}((q, h), \mathcal{N}((q, h))),
\end{align}
where $\tilde{F}$ is a $G$-equivariant function built from a given local feature extraction $F$.
In Eq.~\eqref{eq:group_layer_V3_h}, we need to define what is the local area $\mathcal{N}((q,h))$ around a point $(q,h) \in \mathcal{P}^{(l-1)} \times G$.
Different ways of choosing local neighbors will lead to different architectures.
To simplify the formulation, we choose the local neighbor as
\begin{align*}
\mathcal{N}((q,h)) = \left\{ (p,h) \,|\, p \in \mathcal{N}(q) \right\}.
\end{align*}
Then Eq.~\eqref{eq:group_layer_V3_h} becomes
\begin{align}\label{eq:group_layer_v3}
f^{(l)}_{(q,h)} = \tilde{F}((q, h), \mathcal{N}(q)).
\end{align}

The algorithm for computing the output of $\Phi^{(l)}$ can be separated into three steps as follows:

\begin{algorithm}[G-PointX]\label{alg:GPointNetv3}
\textbf{Input} is a point cloud $\mathcal{X} = \left\{ \left((p,g),f_{p,g}^{(l-1)}\right)  \right\}_{(p,g) \in \mathcal{P}^{(l-1)} \times G}$ and a non-$G$-equivariant SOTA model on point clouds based on a local feature extraction $F$ given in \eqref{eq:local_feature_extraction}. While \textbf{Output} is another point cloud $\mathcal{Y} = \left\{ \left((q,h),f_{q,h}^{(l)}\right)  \right\}_{(q,h) \in \mathcal{P}^{(l)} \times G}$.
\begin{itemize}
	\item[] \textbf{Step 1 (Sampling and grouping).} We determine a subset $\mathcal{P}^{(l)} = \{q_1,\ldots,q_K\}$ of $\mathcal{P}^{(l-1)}$ by using the furthest point sampling algorithm, and then determine the local area $\mathcal{N}(q) = \{p_{1},\ldots,p_{C}\}$ around each point $q=q_j$ by using the nearest point algorithm as: 
	\begin{equation*}
	\mathcal{P}^{(l-1)}
	\xrightarrow{\text{sampling \& grouping}}
	\left\{
		\begin{aligned}
		&q_{1},\mathcal{N}(q_1)\\
		&q_2,\mathcal{N}(q_2)\\
		&\ldots\\
		&q_C,\mathcal{N}(q_C)
		\end{aligned}
	\right.
	\end{equation*}
	
	\item[] \textbf{Step 2 (Local aggregration)}: For each $i$ from 1 to $C$ and for each $h \in G$, we extract a feature vector $f_{q_i,h}^{(l)}$ from each local group $\{(q_i,h),\mathcal{N}((q_i,h))\}$ by
	\begin{align*}
	f'_{q_i,h} = \mathcal{R} \left(  
	\left\{
		F \left( 
		h^{-1}(p-q_i),f_{q_i,h}, f_{p,h}
		\right)
	\right\}_{(p,h) \in \mathcal{N}((q_i,h))}
	\right).
	\end{align*}
%
	\item[] \textbf{Step 3.} Return $\mathcal{Y} = \left\{ ((q_i,h),f'_{q_i,h}) \right\}_{i,h}$
	
%
\end{itemize}
\end{algorithm}

One can verify that
\begin{theorem}
The group layer given in Algorithm~\ref{alg:GPointNetv3} is equivariant with respect to transformations in the semigroup 
 transitions in $\mathbb{R}^3 \rtimes G$.
\end{theorem}

By applying the above algorithm to typical backbones using in practice PointNet++ and PointConv, we obtain $G$-PointNet++ and $G$-PointConv which can be described in detail as follows:
\begin{itemize}
    \item $G$-PointNet++:
        \begin{equation*}
            ((q,h), \, \mathcal{N}(q))
            \rightarrow 
            \left\{
            \begin{aligned}
            h^{-1}(p_1-q), f_{p_1} &\xrightarrow{\text{MLP}_1}
            \text{MLP}_1(h^{-1}(p_1-q), f_{p_1}\oplus f_{p_1})\\
            h^{-1}(p_2-q), f_{p_2} &\xrightarrow{\text{MLP}_1}
            \text{MLP}_1(h^{-1}(p_2-q), f_{p_2}\oplus f_{p_2})\\
            &\quad\ldots\\
            h^{-1}(p_K-q), f_{p_L} &\xrightarrow{\text{MLP}_1}
            \text{MLP}_1(h^{-1}(p_K-q), f_{p_K}\oplus f_{p_K})
            \end{aligned}
            \right\}
            \xrightarrow{\text{Max}}       
            \text{Max}(f_{\mathcal{N}(q)})
            \xrightarrow{\text{MLP}_2}
            f'_q.
        \end{equation*}
    \item $G$-PointConv
        \begin{equation*}
        (q,\, \mathcal{N}(q))
        \rightarrow 
        \left\{
        \begin{array}{lcr}
        \begin{bmatrix}
        ||p_1-q||\\
        \ldots\\
        ||p_{K}-q||
        \end{bmatrix}
        & \xrightarrow{\text{MLP}_1}
        & \begin{bmatrix}
        \text{MLP}_1(||p_1-q||)\\
        \ldots\\
        \text{MLP}_1(||p_{K}-q||)
        \end{bmatrix}
        \\
        \begin{bmatrix}
        h^{-1}(p_1-q)\\
        \ldots\\
        h^{-1}(p_{k}-q)
        \end{bmatrix}
        & \xrightarrow{\text{MLP}_2}
        & \begin{bmatrix}
        \text{MLP}_2(h^{-1}(p_1-q))\\
        \ldots\\
        \text{MLP}_2(h^{-1}(p_K-q))
        \end{bmatrix}
        \\
        \begin{bmatrix}
        f_{p_1}\\
        \ldots\\
        f_{p_K}
        \end{bmatrix}
        & \longrightarrow
        & \begin{bmatrix}
        f_{p_1}\\
        \ldots\\
        f_{p_K}
        \end{bmatrix}
        \end{array}
        \right\}
        \xrightarrow{\odot}
        f_{\mathcal{N}(q)}
        \xrightarrow{\sum}
        f'_q.
        \end{equation*}
\end{itemize}


%
%

\section{Experiments:}
To evaluate the performance of our technique, we run experiments on two tasks which are classification using ModelNet40 dataset and semantic segmentation using S3DIS dataset. In particular, we initially compare our methods with other equivariant models on $SO(3)$ rotated dataset to show the enhancement on both performance and complexity. Thereafter we do the ablation study which highlights the benefits of using our techniques, compared to solely employ rotation augmentation on original models. Additionally, the ablation study also provides the information about the performance of different group $G$. Eventually, we conduct experiments  on semantic segmentation using the G equivariant models and the original ones. 

\subsection{GPointX versus other equivariant models}
In this section we compares the performance of $G$-PointNet++ and $G$-PointConv with other equivariant models on $SO(3)$ rotated ModelNet 40 dataset, which contains 13,834 mesh samples from 40 labels such as: table, chair, plane, plant, etc. Notably, the $G$ group includes 24 rotation angles which consists of the combination of any $\pm \frac{\pi}{2}$ and $\pm \pi$ rotations around the three axes. In terms of the experiment configuration, we reuse the same training pipeline of original models; number of points and number of epochs are 1024 and 200, respectively. However, to keep the memory required the same as original model, batch size is reduced to 8. Moreover, we also add $SO(3)$ rotation along with augmentations employed in the original paper. The results in Table~\ref{table:rotation} indicates that both $G$PointNet++ and $G$PointConv are outperforms other models in the literature.
\begin{table}[h]
\begin{center}
\begin{tabular}{ |c|c|c|c|c|c| } 
\hline
Name & $SO(3)$ Rotated Dataset \\
\hline
QENet \cite{zhao2020quaternion} &74.4\\
Model in \cite{zhang2019rotation} &86.5\\
ClusterNet \cite{chen2019clusternet} &87.1\\
SPConv \cite{Chen2021CVPR} &88.3\\
Model in \cite{li2020rotation} &89.4\\
\hline
G24PointNet++ (Augmentation + $SO(3)$) & \textbf{90.3}\\
G24PointConv (Augmentation + $SO(3)$) & \underline{89.6}\\
\hline
\end{tabular}
\end{center}
\caption{\label{tab:table-name} Results of different equivariant model on $SO(3)$ Rotated ModelNet40.}
\label{table:rotation}
\end{table}

\begin{figure}[h]
\begin{center}
\includegraphics[width = \textwidth] {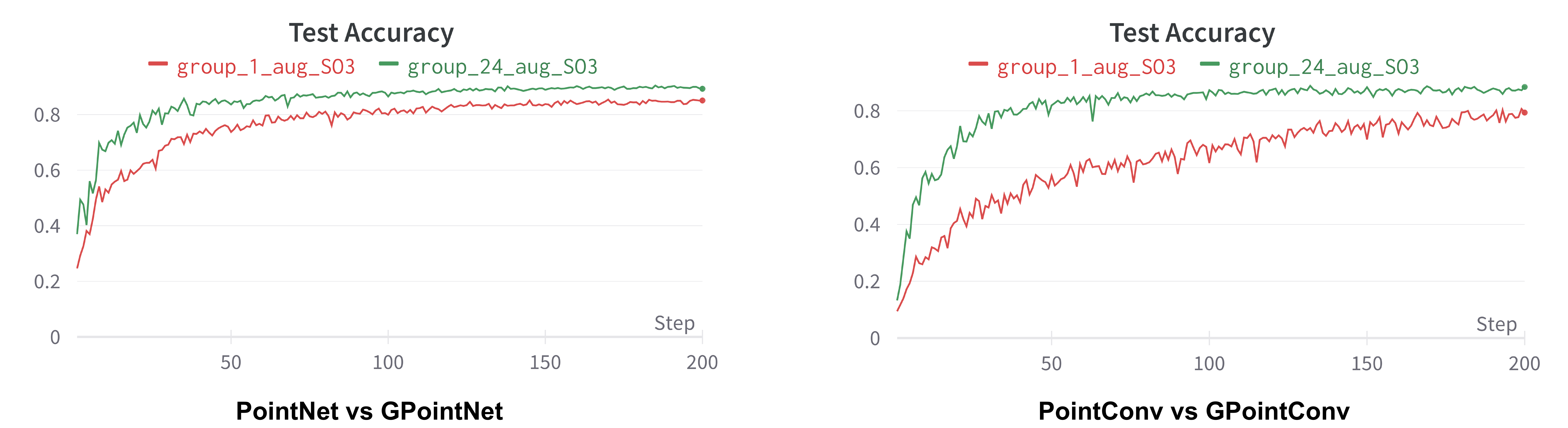}
\caption{Convergence speed of the test accuracy on the original test set of $G$-PointConv and $G$-PointNet++ using $G=G_{24}$ in comparison with standard models PointConv and PointNet++ with data augmentation using the same sampled rotations.}
\label{figure:classification}
\end{center}
\end{figure}

\subsection{Ablation study on different groups}
We compare the accuracy of $G$-PointNet++ and $G$-PointConv  for object classification on ModelNet40 when using different finite rotation groups $G_1$, $G_4$, $G_{12}$ and $G_{24}$.
Here, $G_n$ is a subgroup of SO(3) with $n$ rotations.
In particular, $G_4$ contains the combinations of the $\frac{\pi}{2}$ rotations around the $z$-axis.
$G_{12}$ contains the combination of the $\frac{\pi}{3}$ rotations around the $z$-axis and the $\pi$ rotations around the $y$-axis.
Note that, when using $G_1$, the Group Equivariant models are equivalent to original models. Table~\ref{table:ablation} shows that when we increase the elements in group $G$ in the results of the model on $SO(3)$ rotated dataset is also improved. Furthermore, this experiment also highlights the large enhancement of our group equivariant technique, compared to using solely rotation augmentation. To fairly compare the effect of the different groups, we fixed the batch size of all experiments equals to 8 and regarding the remaining parameters, we use the same configurations with the previous section.

\begin{table}[h]
\begin{center}
\begin{tabular}{ |c|c|c|c|c|c| } 
\hline
Name & $G_1$ & $G_4$ & $G_{12}$ & $G_{24}$ \\
\hline
$G$PointNet++&86.00&88.0&89.6&\textbf{90.3}\\
$G$PointConv&80.4&84.6&88.3&\textbf{89.6}\\
\hline
\end{tabular}
\end{center}
\caption{\label{tab:table-name} Results of different groups on $SO(3)$ Rotated ModelNet40.}
\label{table:ablation}
\end{table}

\subsection{Semantic Segmentation on S3DIS}
Regarding semantic segmentation, S3DIS dataset was used to evaluate the performance of group equivariant and original models. This dataset contains contains 271 rooms and the objects are divided into 13 classes. Here, $G$ is set to 8 instead of 24 since the objects in the rooms were mostly rotated around Oz. Similar to the previous section, we also fixed the batch size, epochs and number of points of the three models, which are 16, 32, and 4096 and the remaining parameter were kept as the same as the papers. As observed from Table~\ref{table:segmentation}, there is a significant improvement in terms of performance of PointConv\cite{wu2019pointconv} when using our equivariant method.

\begin{table}[h]
\begin{center}
\begin{tabular}{ |c|c|c|c|c|c| } 
\hline
Name & Original Model & Group Equivariant Version \\
\hline
PointNet++ & 0.535 & \textbf{0.546}  \\
PointConv & 0.530 & \textbf{0.578}\\
\hline
\end{tabular}
\end{center}
\caption{\label{tab:table-name} Results of PointX and GPointX on S3DIS.}
\label{table:segmentation}
\end{table}

\begin{figure}[h]
\begin{center}
\includegraphics[scale=0.3]{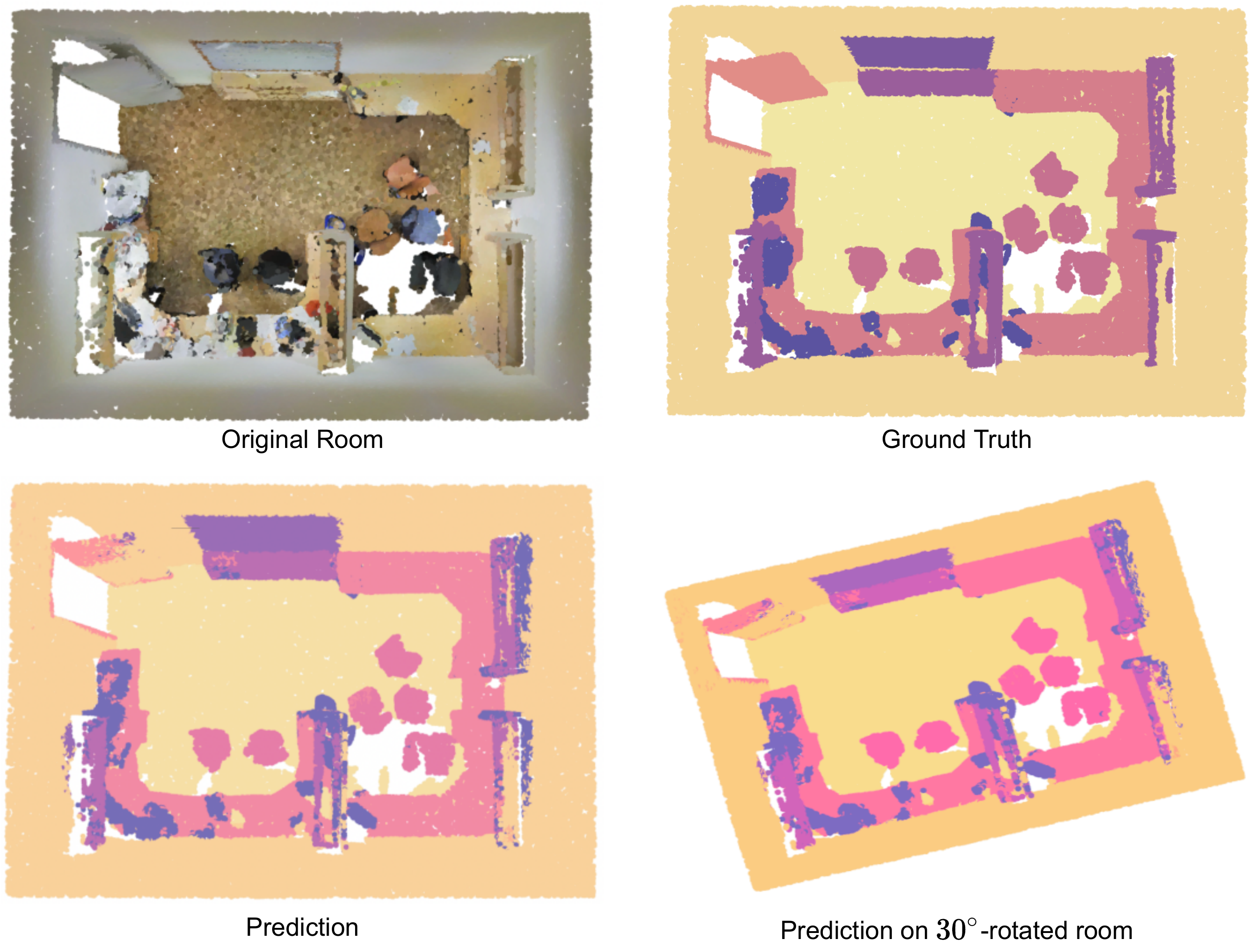}
\caption{Semantic segmentation with S3DIS dataset.}
\label{figure:segmentation}
\end{center}
\end{figure}





\end{document}